\documentclass{article}
\usepackage{preprint,times}

\usepackage{hyperref}
\usepackage{url}

\usepackage{graphicx}
\usepackage{booktabs}
\usepackage{multirow}
\usepackage{float}
\usepackage{comment}
 \usepackage{amsmath}
\usepackage{amssymb}
\usepackage{mathtools}
\usepackage{amsthm}
\usepackage{subcaption}
\usepackage[capitalize,noabbrev]{cleveref}
\hypersetup{
    colorlinks=true,
    linkcolor=black,
    citecolor=black,
    urlcolor=black
}

\title{Puffin: Probabilistic Learning of Spatial\\Detail From Coarse Observations}

\author{%
  Chaitanya Jobanputra \\
  DFKI\thanks{German Research Center for Artificial Intelligence.} \\
  Kaiserslautern, Germany
  \And
  Sebastian Vollmer \\
  DFKI\footnotemark[1] \\
  Kaiserslautern, Germany
  \And
  Gerrit Gro{\ss}mann\thanks{Correspondence to \texttt{Gerrit.Grossmann@dfki.de}.} \\
  DFKI\footnotemark[1] \\
  Kaiserslautern, Germany
}

\finalcopy
\begin{document}

\maketitle
\lhead{}
\fancyfoot[L]{\footnotesize Preprint. Under review.}

\begin{abstract}
High-resolution socioeconomic variables are important for applications such as urban planning, public health, disaster response, and resource allocation. In practice, however, these variables are often observed only at a coarse spatial resolution. We introduce \textsc{Puffin}, a probabilistic framework for statistical disaggregation that raises the resolution of coarse totals using high-resolution satellite embeddings as covariates. Instead of predicting a single value for each fine-resolution subregion, \textsc{Puffin} learns a probability distribution and is trained through an aggregation-aware likelihood. At inference, \textsc{Puffin} conditions these predictions on the observed regional total and splits it among the subregions. The resulting fine-scale estimates are consistent with the observed aggregate and come with calibrated uncertainty, without requiring fine-resolution labels for training. We evaluate \textsc{Puffin} on German and US census, employment, and election data across population, jobs, and other count variables, and study when statistical disaggregation succeeds or fails across regions, countries, and targets. Code is available at 
\href{https://github.com/gerritgr/Puffin}{\small\texttt{github.com/gerritgr/Puffin}}.
\end{abstract}

\section{Introduction}
\begin{figure*}[h!]
    \centering
    \includegraphics[width=0.92\linewidth]{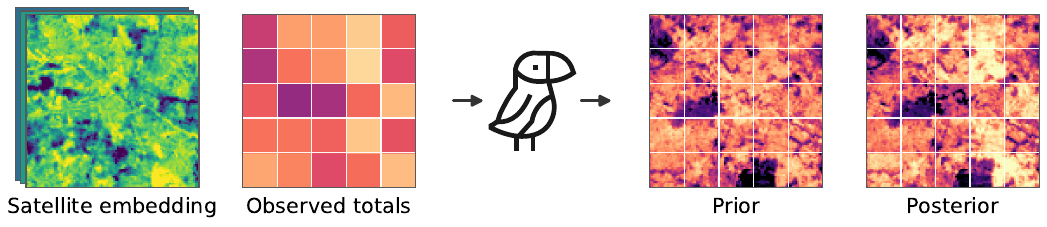}
    \caption{
    \textbf{Disaggregation as probabilistic constraint satisfaction.} From a satellite embedding and one observed total per region, \textsc{Puffin} learns a prior distribution over every subregion and conditions it on the totals, which gives a posterior that splits each total exactly among its subregions (means shown).
    }
    \label{fig:motivating_example}
\end{figure*}

Where people live, where they work, and where they are exposed to risk are questions that urban planning, public health, and disaster response must answer street by street. Censuses and registers publish such statistics only as one total per reporting unit, a census tract or a county, to protect the people behind the numbers and because surveying at a finer scale is expensive. Satellite imagery, by contrast, covers the whole world at $10$ to $100\,$m and carries a strong signal about socioeconomic outcomes \citep{jean2016combining,yeh2020using,rolf2021generalizable}, and recent foundation models such as AlphaEarth turn every pixel into a short vector, an embedding \citep{brown2025alphaearth}, building on self-supervised pretraining for satellite imagery \citep{cong2022satmae}. Recovering the fine-scale map behind a coarse total from such imagery is called spatial disaggregation, or areal interpolation \citep{flowerdew1991using,metzger2022fine}, and it is the problem we study.

We think of disaggregation as splitting a fixed amount. Each region reports one total, which we divide among the small subregions inside it so that every subregion gets a non-negative value and the pieces add back up to the reported total.
This mass-preserving view goes back to Tobler's interpolation \citep{tobler1979smooth}, and we make the split learned and probabilistic. The only supervision is that single total, which makes this the count version of learning from label proportions \citep{quadrianto2008estimating}.

The gridded population products in widest use today, among them WorldPop \citep{stevens2015disaggregating}, the Global Human Settlement population grid (GHS-POP) \citep{Schiavina_GHS-POP_R2023A_2023}, and LandScan \citep{lebakula2025landscan}, are still built by dasymetric mapping \citep{eicher2001dasymetric,mennis2003generating}, which spreads each regional total over auxiliary layers assumed to indicate where people are, such as night lights, roads, or building footprints, weighted by hand or fitted to the totals. Those layers are laborious to assemble and differ from country to country, and the published product is a single number per cell. 

We introduce \textsc{Puffin}, a probabilistic framework for spatial disaggregation that removes both limitations. It replaces the curated layer stack with a single satellite embedding and returns a predictive distribution per subregion rather than a bare number, while still reproducing each published total exactly. Training needs no fine-resolution labels, which is what makes the point estimates usable where such labels do not exist.

\textsc{Puffin} treats disaggregation as probabilistic constraint satisfaction (Figure~\ref{fig:motivating_example}). A small network predicts a prior distribution over each subregion's count from its embedding, trained only through the aggregation constraint, and conditioning that prior on an observed total gives a posterior that satisfies the constraint exactly. Its only inputs are a regional total and a globally available satellite embedding, so it applies wherever totals are published, including the many countries with no fine-resolution census grid.

\paragraph{Contributions.}
\textsc{Puffin} frames spatial disaggregation as learning from aggregates under an exact aggregation constraint, and from this we make three contributions.
\begin{itemize}
\item \textbf{Disaggregation as Probabilistic Constraint Satisfaction.} We learn a per-subregion predictive distribution from regional totals alone, with the aggregation constraint as the only supervision, and cast inference as conditioning that distribution on the observed totals. No fine-resolution label is needed to train the model.
\item \textbf{Mass Conservation With Calibrated Uncertainty.} The resulting posterior is closed-form and consistent with every aggregate. The estimate it yields is non-negative, reproduces each observed total exactly, and gives each subregion a share of the correction set by its uncertainty. A two-stage calibration, in which only one factor per setting uses fine labels, turns the posterior variances into intervals that hold their nominal marginal coverage across resolutions, countries and targets.
\item \textbf{A Held-Out Region Evaluation.} On regions held out by spatial cross-validation and against three baselines, we evaluate the approach at many resolutions, on a city (Berlin) and a whole country (Germany), on a second country with irregular units (the United States), and on two variables (population and jobs). We also identify where a linear baseline is as good and where the approach does not help (precinct turnout and rare events).
\end{itemize}

\section{Notation and Problem Setting}
The study area consists of $N$ coarse regions, each divided into $M$ subregions, and $(i,j)$ denotes subregion $j$ of region $i$. Regions are the level at which a total is observed (e.g., census tracts or counties), and subregions are the level we want to recover (grid cells in Germany, census blocks in the United States). For notational simplicity we assume every region has the same number $M$ of subregions, all of the same size. Neither assumption is needed in practice, and both are dropped in Section~\ref{sec:impl}.

The unobserved fine outcomes form a matrix $Y\in\mathbb{R}_{\ge0}^{N\times M}$ whose entry $y_{i,j}$ is the outcome of subregion $j$ of region $i$, for instance the number of residents of that grid cell or census block. The observed coarse outcomes form a vector $\widehat{Y}\in\mathbb{R}_{\ge0}^{N}$, where the hat marks the aggregated level rather than an estimate, whose entry $\widehat{Y}_i$ is the single observable
total of region $i$, for instance the number of residents of that census tract. Each subregion carries contextual covariates, collected in a tensor $C\in\mathbb{R}^{N\times M\times d}$ whose entries $\mathbf{c}_{i,j}\in\mathbb{R}^{d}$ are $d$-dimensional feature embeddings of the subregion (e.g., $64$-dimensional AlphaEarth embeddings); nothing below depends on this choice. Fine and coarse outcomes are linked by sum aggregation,

\begin{equation}
\widehat{Y}_i=\sum_{j=1}^{M} y_{i,j}\quad\text{for } i=1,\dots,N,
\label{eq:agg}
\end{equation}
so a single number per region is shared by all of its subregions, while $Y$ itself is never observed. This is a weakly supervised, learning-from-aggregates setting.

\paragraph{Objective.}
Given the coarse totals $\widehat{Y}$ and the covariates $C$, our goal is to infer $Y$, consistently with the aggregates of Equation~\ref{eq:agg} and with uncertainty over the inferred values.

\section{Our Method: \textsc{Puffin}}

\textsc{Puffin} works in two stages. It learns a prior over every subregion's outcome from the covariates, fitted only through the aggregation constraint, and then conditions that prior on the observed total, which gives a closed-form posterior that conserves the total exactly.

\subsection{Overall Idea}

For each subregion $j$ of region $i$, a neural network takes the context $\mathbf{c}_{i,j}$ as input and predicts the mean and variance $(\mu_{i,j},\sigma_{i,j}^2)=f_{\boldsymbol{\theta}}(\mathbf{c}_{i,j})$ of a Gaussian distribution over the target variable.

\paragraph{Prior.}
We interpret the joint distribution induced by these predictions as a
\emph{prior} over the fine-resolution outcomes. In particular,
\begin{equation}
\tilde y_{i,j}
\sim
\mathcal{N}(\mu_{i,j},\sigma_{i,j}^2),
\qquad j=1,\ldots,M,
\label{eq:prior}
\end{equation}
with the subregions assumed to be independent.
Here, $\tilde y_{i,j}$ denotes the model's random variable for the unobserved
ground-truth outcome $y_{i,j}$.
This prior represents what the model expects for each subregion based only on
its covariates, before taking the observed regional aggregate into account.

\paragraph{Posterior.}
For each region, we observe only the aggregate
$
\widehat{Y}_i
=
\sum_{j=1}^{M} y_{i,j}.
$
We obtain the \emph{posterior} distribution over the model's subregion outcomes
by conditioning the prior on this observed aggregate:
\[
p_{\boldsymbol{\theta}}\left(
\tilde y_{i,1},\ldots,\tilde y_{i,M}
\;\middle|\;
\sum_{j=1}^{M} \tilde y_{i,j}=\widehat{Y}_i
\right).
\]

For the Gaussian prior, the posterior is again Gaussian (restricted to the aggregation constraint), and its maximum a posteriori (MAP) estimate is equal to its mean and available in closed form (Equation~\ref{eq:post} in Section~\ref{sec:impl}).
Intuitively, if the observed aggregate is larger than the sum of the predicted means, the individual predictions must be corrected upward, and if it is smaller, they must be corrected downward. The correction is distributed across subregions according to their uncertainty, such that subregions with larger variance are adjusted more.

\paragraph{Training.}
Because independent Gaussian variables add to another Gaussian, the model's prior over the regional aggregate is again Gaussian, with mean $\mu_i=\sum_{j}\mu_{i,j}$ and variance $\sigma_i^2=\sum_{j}\sigma_{i,j}^2$.
We train the neural network through the prior only, never through the posterior, by maximizing the likelihood that the prior assigns to the observed regional aggregates, or equivalently by minimizing the negative log-likelihood (NLL)
\begin{equation}
\mathcal{L}(\boldsymbol{\theta})
=
\frac{1}{|\mathcal{T}|}
\sum_{i\in\mathcal{T}}
\left[
\frac{
\left(
\widehat{Y}_i-\sum_{j=1}^{M}\mu_{i,j}
\right)^2
}{
\sum_{j=1}^{M}\sigma_{i,j}^2
}
+
\log\left(
\sum_{j=1}^{M}\sigma_{i,j}^2
\right)
\right],
\label{eq:nll}
\end{equation}
over the set $\mathcal{T}$ of training regions, up to an additive constant and an overall positive factor. It reduces to the squared loss used by CLAM \citep{grossmann2026clam}, a related disaggregation model, when all variances are equal and every region has the same number of subregions.
The prior does not take the total as input, so its mean may well miss the observed $\widehat{Y}_i$. That residual is exactly what the loss penalizes, and reproducing a total exactly is the task of the posterior, not of the prior. A different additive variable needs only its own aggregate likelihood in place of this one.

\subsection{Implementation}
\label{sec:impl}

Four things separate the idea above from a working method. Subregions differ in area, the split has to stay non-negative, the posterior variances have to be turned into trustworthy intervals, and the function $f_{\boldsymbol{\theta}}$ needs a concrete form.

\paragraph{Uneven Subregion Sizes.} 
Real subregions are not equally large, and a large block should be able to hold more than a small one that looks the same. We assume an area matrix $A\in\mathbb{R}_{>0}^{N\times M}$, where $a_{i,j}$ contains the area of subregion $(i,j)$, measured in units of $100\,$m cells. The network then predicts a mean and a variance per unit area, and the area turns both into a count,
\begin{equation}
(u_{i,j},\,v_{i,j})=f_{\boldsymbol{\theta}}(\mathbf{c}_{i,j}),\qquad
\mu_{i,j}=a_{i,j}\,u_{i,j},\qquad
\sigma_{i,j}^2=a_{i,j}\big(\mathrm{softplus}(v_{i,j})+\varepsilon\big),
\label{eq:area}
\end{equation}
with $\varepsilon=10^{-6}$ keeping the variance positive. Everything above is unchanged, because prior, loss, and posterior only ever use $\mu_{i,j}$ and $\sigma_{i,j}^2$. On a uniform grid the area is constant and cancels. On US census blocks, whose areas span about four orders of magnitude between the $1$st and $99$th percentile, it is what makes the split work. Scaling the variance with the area lets a large block absorb a correspondingly larger share of the region's correction. Dropping that factor while keeping everything else fixed costs about $0.10$ within-region $R^2$ on the US blocks (Appendix~\ref{app:details}). A varying number of subregions per region needs no change at all, beyond reading $M$ as $M_i$.

\paragraph{Closed-Form Split and Non-Negative Estimates.} Conditioning a Gaussian on the value of a sum has a closed form, so the split needs no optimization at inference,
\begin{equation}
\tilde y_{i,j}^{\mathrm{MAP}} = \mu_{i,j} + \frac{\sigma_{i,j}^2}{\sum_{k=1}^{M}\sigma_{i,k}^2} \left( \widehat{Y}_i-\sum_{k=1}^{M}\mu_{i,k} \right).
\label{eq:post}
\end{equation}
Counts cannot be negative, but this estimate can be, wherever the prior over-predicts a region. We clip negative $\tilde y_{i,j}^{\mathrm{MAP}}$ to zero and rescale the region so that the estimates still sum to $\widehat{Y}_i$, falling back to an even split if all of a region's estimates vanish. We write $\tilde{\mu}_{i,j}$ for the resulting value and call it the \emph{\textsc{Puffin} estimate}. It is the posterior mean after clipping, it is our estimate of $y_{i,j}$, and it is what every map and table below reports, while the posterior variances give the intervals.

\paragraph{Calibrated Intervals.} Conditioning on the total also gives each subregion a posterior variance, $\tilde{\sigma}_{i,j}^2=\sigma_{i,j}^2-\sigma_{i,j}^4/\sigma_i^2$ with $\sigma_i^2=\sum_k\sigma_{i,k}^2$. They come from a prior fitted to totals alone and are smaller than the errors the estimate actually makes, so we rescale them in two steps. Within each $10\,$km tile $t$ we first use the coarse residuals, which need no fine labels, to set $\gamma_t^2=\sum_{i\in t}(\widehat{Y}_i-\mu_i)^2/\sum_{i\in t}\sigma_i^2$ and $\bar{\sigma}_{i,j}^2=\gamma_{t(i)}^2\,\tilde{\sigma}_{i,j}^2$. On a set of validation regions we then choose a single factor $\kappa$ as the $90\%$ quantile of the standardized errors $|y_{i,j}-\tilde{\mu}_{i,j}|/\bar{\sigma}_{i,j}$, which gives the interval $\tilde{\mu}_{i,j}\pm\kappa\,\bar{\sigma}_{i,j}$. Only $\kappa$ uses fine labels, and only on validation regions, which are always disjoint from the regions we evaluate on (Section~\ref{sec:experiments}). Appendix~\ref{app:details} gives what a tile is and which totals each stage uses.

\paragraph{Network and Optimization.} The network $f_{\boldsymbol{\theta}}$ is a single hidden layer of width $32$ with Gaussian error linear unit (GELU) activations \citep{hendrycks2016gaussian} and about $2{,}000$ parameters. We minimize Equation~\ref{eq:nll} as written, standardizing the totals by the average training density so the loss does not depend on the scale of the variable, and, for the German country-scale configurations that do not fit in memory, sampling whole regions into minibatches, since a region's total needs all of its subregions in the batch. Appendix~\ref{app:details} gives the remaining settings.

\section{Experiments}
\label{sec:experiments}

We run three experiments on real census and election data, where the fine-scale values are counted rather than modeled, so unlike gridded population products they can validate a disaggregation. Both offices do perturb the published counts slightly for disclosure control \citep{abowd20222020,destatis2024cellkey}, which puts a floor under the error any method can reach. We first give the protocol and the baselines shared by all three.

\paragraph{Prior Versus Posterior.} The prior (Equation~\ref{eq:prior}) is what one would use where no total is available, and it enters the experiments only through the training loss. Since every region's total is observed here, every reported number uses the posterior (Equation~\ref{eq:post}), through the \textsc{Puffin} estimate for point accuracy and the posterior variances for the intervals. The exceptions are the \textsc{Puffin} (prior) rows, which score the prior alone against its own regional mean.

\paragraph{Metric.} We compare the \textsc{Puffin} estimate $\tilde{\mu}_{i,j}$ with held-out true values using within-region $R^2$, which removes each region's mean $\bar{y}_i=\widehat{Y}_i/M$ from both truth and estimate,
\begin{equation}
R^2_{\mathrm{within}}=1-\frac{\sum_{i,j}\big(\dot{y}_{i,j}-\dot{\mu}_{i,j}\big)^2}{\sum_{i,j}\big(\dot{y}_{i,j}-\overline{\dot{y}}\big)^2},\qquad \dot{y}_{i,j}=y_{i,j}-\bar{y}_i,\quad \dot{\mu}_{i,j}=\tilde{\mu}_{i,j}-\bar{y}_i,
\label{eq:within}
\end{equation}
so the region-average baseline scores exactly zero and the metric measures only how well each total is split among its subregions. For comparison with other work, Appendix~\ref{app:results} also reports subregion-$R^2$, the usual $R^2$ against the global mean. It largely measures the totals rather than the split, especially for coarse subregions, because the observed totals alone already separate dense regions from empty ones. Even the region average reaches $0.83$ on the coarsest Berlin configuration, $10\,$km regions with $5\,$km subregions, against $0.25$ at $10\,$km with $100\,$m subregions (Table~\ref{tab:berlin_cell}).

\paragraph{Protocol.} All results use $5$-fold spatial cross-validation. Regions are grouped into $5\,$km spatial groups that are assigned to folds as a whole, so that neighboring regions share a fold. Where a region is larger than the group, as in the $10\,$km grid, for nearly all counties, and for about a quarter of US tracts, each region forms its own group and the assignment reduces to a random split per region, so a test region can border a training one. Since training uses only totals, one would in practice fit on all regions, so this is a stricter test than deployment requires. For each test fold the preceding fold is used for validation (checkpoint selection, the ridge strength of the regression baseline below, and $\kappa$) and the remaining three for training, so every subregion is scored once, out of fold. We report within-region $R^2$ (Equation~\ref{eq:within}) on all subregions with the coverage of the $90\%$ intervals, and Appendix~\ref{app:results} adds subregion-$R^2$ and the inhabited-only comparisons. \textsc{Puffin} results are $5$-seed means with fixed folds, so the reported standard deviations cover the initialization, not the choice of fold. Irregular units use the area term of Equation~\ref{eq:area} for every method, except precinct turnout, where precincts hold comparable numbers of voters so the area actively misleads and no method uses it.

\paragraph{Baselines.} We compare \textsc{Puffin} with three baselines, all using the same regions, subregions, and folds, never using fine labels, and reproducing every observed total exactly, so the comparison isolates how each method splits a total. The \textbf{region average} gives every subregion the same share $\widehat{Y}_i/M$ and defines the zero point of within-region $R^2$. \textbf{Bicubic interpolation} reads the regional densities as samples of a smooth surface and evaluates a cubic interpolant of it at every subregion, so it tests whether spatial smoothness alone explains the fine structure, using neighboring totals but no covariates. \textbf{Goodman-style ecological regression} \citep{goodman1953ecological,goodman1959some} infers individual-level relationships from aggregate data by regressing region-level rates on region-level covariates, which we apply to the same embeddings as \textsc{Puffin}, so it tests whether a nonlinear probabilistic model is needed at all. Both fitted densities are multiplied by the subregion area, clipped at zero, and rescaled to the observed total. The baselines are deterministic and therefore have no seeds. Appendix~\ref{app:details} gives their exact form.

\begin{figure}[t]
\centering
\includegraphics[width=1.0\linewidth]{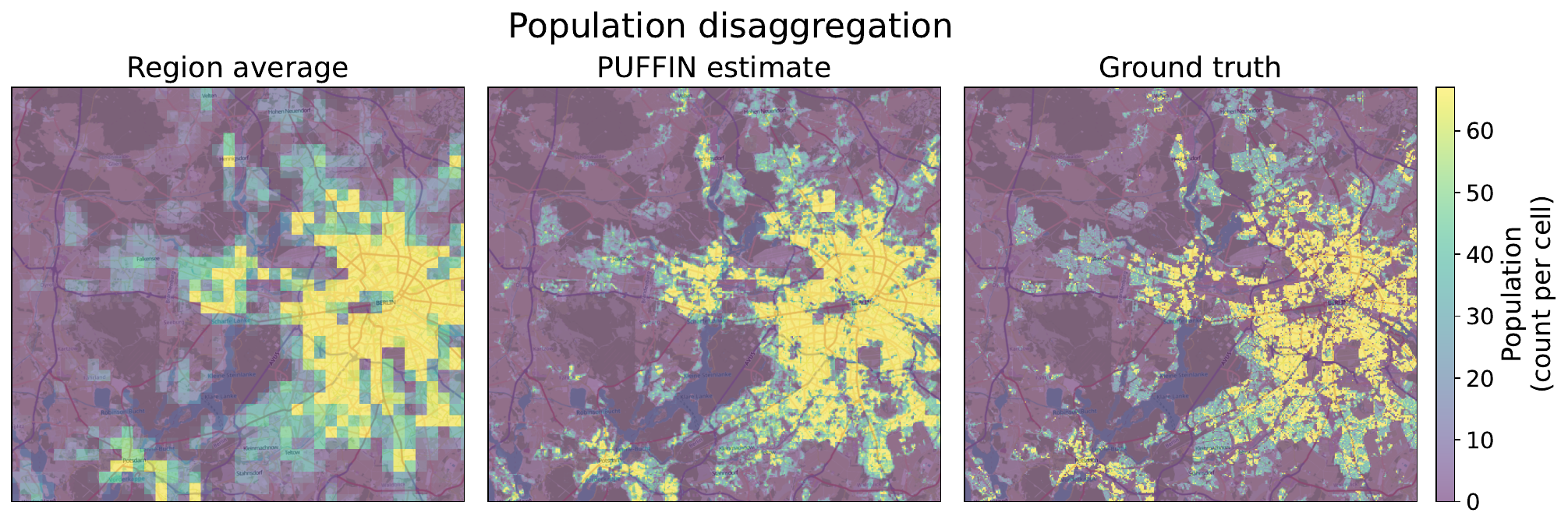}
\caption{\textbf{Qualitative disaggregation (Berlin, $1\,$km regions split into $100\,$m subregions).} Region-average baseline (left), \textsc{Puffin} estimate (middle), and ground truth (right), for population. \textsc{Puffin} recovers within-region structure that the flat baseline misses, from the embeddings and totals alone.}
\label{fig:berlin_map}
\end{figure}

\subsection{Experiment 1: German Census, City to Country}

\paragraph{Data.} The German register-based Zensus 2022 counts residents on a $100\,$m grid \citep{Zensus2022}, so the fine values are measured rather than modeled. We use every $100\,$m cell in a $102\times96\,$km box around Berlin ($979{,}200$ cells, referred to as Berlin below) and in Germany ($35.7$M), and form nested square grids. The covariates are AlphaEarth embeddings averaged over each subregion. The regions are the coarse squares whose totals we observe, with sizes from $200\,$m to $10\,$km, and the subregions are the finer squares inside them, $2$ to $100$ per side, which gives $18$ configurations. Appendix~\ref{app:details} lists the cell counts and the configurations.

\paragraph{Within a City (Berlin).} Berlin tests the core question in the simplest setting, namely whether, within a single city and its surroundings, the embeddings can tell which parts of a region hold its people. The uniform grid removes the area term, and varying the region and subregion sizes shows how the answer depends on scale. \textsc{Puffin} answers yes in every one of the $18$ configurations (Table~\ref{tab:within_main}, top), where the region average scores zero by construction, and it does better the larger the subregion, from $0.12$ at $200\,$m regions with $100\,$m subregions to $0.68$ at $10\,$km regions with $5\,$km subregions. Bicubic interpolation reaches only $0.142$ on average, so the totals of neighboring regions alone do not reveal the fine structure, pointing to the embeddings as the source of the gain. Goodman-style regression keeps up on a single city and is slightly ahead on average ($0.448$ versus $0.427$), winning $12$ of the $18$ configurations, but its lead comes from the largest regions and the coarsest subregions. The $10\,$km row of the grid holds six configurations, one per subregion size, and Berlin has only $73$ regions of that size to train on. Goodman's ridge-regularized linear model wins all six of them, and the fine-scale error of \textsc{Puffin} starts to rise after about $100$ of the $2{,}000$ epochs (Figure~\ref{fig:lc_main}), which is what overfitting so few totals looks like. Germany has $2{,}317$ regions of the same size, and there \textsc{Puffin} wins all six. Over the remaining twelve configurations the two are close in Berlin, $0.008$ apart on average. Figure~\ref{fig:berlin_map} shows the disaggregation on a map.

\paragraph{Scaling to a Country (Germany).} 
Next, we train on all of Germany, which provides about $36\times$ more data than Berlin. With this much data, \textsc{Puffin} reaches within-region $R^2$ between $0.18$ and $0.84$ (Table~\ref{tab:within_main}), higher than on Berlin in every configuration, and it now beats Goodman-style regression in $16$ of $18$ configurations (mean $0.561$ versus $0.544$). The margins are small, at most $0.04$, but consistent. The prior alone stays below Goodman ($0.483$ versus $0.544$), which suggests that what puts \textsc{Puffin} ahead is the variance-weighted split rather than the nonlinearity of the prior. Bicubic interpolation again stays far below, at $0.100$ on average. The per-configuration tables for both datasets, the inhabited-only cells, and subregion-$R^2$ are in Appendix~\ref{app:results}.

\begin{table}[t]
\centering\small
\setlength{\tabcolsep}{4pt}
\begin{tabular}{llcccccc}
\toprule
& & \multicolumn{6}{c}{Subregions per side ($R/s$)} \\
\cmidrule(lr){3-8}
Data & Region size $R$ & $2$ & $5$ & $10$ & $20$ & $50$ & $100$ \\
\midrule
\multirow{5}{*}{Berlin} & $200\,$m & .12 & -- & -- & -- & -- & -- \\
 & $1\,$km & .41 & .38 & .28 & -- & -- & -- \\
 & $2\,$km & .49 & .48 & .42 & .30 & -- & -- \\
 & $4\,$km & .54 & .51 & .47 & .39 & -- & -- \\
 & $10\,$km & .68 & .58 & .52 & .45 & .38 & .28 \\
\midrule
\multirow{5}{*}{Germany} & $200\,$m & .18 & -- & -- & -- & -- & -- \\
 & $1\,$km & .58 & .48 & .35 & -- & -- & -- \\
 & $2\,$km & .70 & .61 & .51 & .37 & -- & -- \\
 & $4\,$km & .77 & .69 & .61 & .50 & -- & -- \\
 & $10\,$km & .84 & .75 & .70 & .62 & .48 & .36 \\
\bottomrule
\end{tabular}

\vspace{6pt}

\begin{tabular}{lccccc}
\toprule
Mean over the $18$ configurations & Reg.-avg. & Bicubic & Goodman & \textsc{Puffin} (prior) & \textsc{Puffin} \\
\midrule
Berlin & .000 & .142 & \textbf{.448} & .341 & .427 \\
Germany & .000 & .100 & .544 & .483 & \textbf{.561} \\
\bottomrule
\end{tabular}
\caption{Within-region $R^2$ on the German census, all $100\,$m cells. \textbf{Top}: \textsc{Puffin} per configuration, a row being the region size $R$ and a column the number of subregions per side, with blanks where subregions would be smaller than $100\,$m. \textbf{Bottom}: the mean over the $18$ configurations per method, best in bold. \mbox{Reg.-avg.} is the region-average baseline, which scores $0$ by construction, and \textsc{Puffin} (prior) is the prior before conditioning, which the posterior improves on everywhere. The prior keeps its own split, whereas the baselines are rescaled to the observed total, so it is a conservative reference rather than a like-for-like one. Tables~\ref{tab:berlin_within} and~\ref{tab:germany} give every method per configuration. \textsc{Puffin}: $5$-seed mean (std $\le 0.024$).}
\label{tab:within_main}
\end{table}

\subsection{Experiment 2: US Census, Population and Jobs}

\paragraph{Data.} The US 2020 census reports population per census block; we retain the $8.0$M blocks of the $48$ contiguous states and the District of Columbia \citep{uscensus2020tiger}, holding $325.1$M residents (Appendix~\ref{app:details} says which blocks are missing and why). These are irregular polygons nested in $83{,}774$ census tracts, and the Longitudinal Employer-Household Dynamics Origin-Destination Employment Statistics (LODES, version~8, the most recent year available per state between $2020$ and $2022$) report the number of jobs located in each block \citep{census_lodes}. The regions are census tracts, whose totals we observe, and the subregions are the census blocks inside them. The covariates are AlphaEarth embeddings averaged over each block polygon, and the area term uses each block's area, whose range is given in Section~\ref{sec:impl}.

\begin{figure}[t]
\centering
\includegraphics[width=1.0\linewidth]{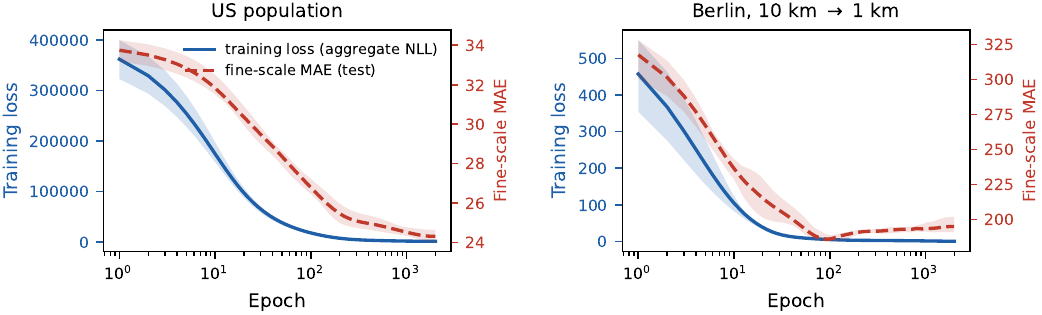}
\caption{\textbf{Training on totals improves the fine-scale split.} Training loss (blue, left axis) and the mean absolute error (MAE) of the \textsc{Puffin} estimate on held-out test subregions (red, right axis) over training, as mean and range over $5$ seeds, for US population (left) and Berlin at $10\,$km (right). No subregion value ever enters training, yet the fine-scale error falls with the aggregate loss. On Berlin at $10\,$km, where only $73$ regions are available for training, it starts to rise again after about $100$ epochs, while the validation totals keep improving until epoch $485$, so selecting the checkpoint on them does not catch this.}
\label{fig:lc_main}
\end{figure}

\paragraph{A Second Country and a Second Variable (US).} This experiment changes two things at once. The fine units are irregular blocks instead of grid cells, and the variables include jobs, which concentrate in a few small commercial blocks instead of following where people live. \textsc{Puffin} works here too (Table~\ref{tab:us}), with population at $+0.377$, inside the range seen in Germany, and jobs at $+0.209$. Here bicubic interpolation and Goodman-style regression both fail. On population they reach $-0.41$ and $-1.04$, worse than the region average, and on jobs only $+0.06$ and $+0.05$. The likely reason is that a smooth or linear density cannot fall to near zero on large empty blocks, so multiplying it by area pushes each tract's total into its largest blocks. The prior alone does not escape this either, at $-0.56$ on population and $-0.05$ on jobs, so as in Germany it is the conditioning step and not the prior that carries the result: scaling the variance with the area lets a large block take a larger share of its region's correction. Figure~\ref{fig:lc_main} shows that this skill is learned from the totals alone. Subregion-$R^2$ and inhabited blocks leave \textsc{Puffin} ahead on both targets, with the two baselines swapping on jobs (Tables~\ref{tab:us_cell} and~\ref{tab:us_inhab} in Appendix~\ref{app:results}). Figure~\ref{fig:usmap} shows population in New York City, and Figure~\ref{fig:usjobs} in Appendix~\ref{app:results} shows jobs.

\begin{figure}[t]
\centering
\includegraphics[width=1.0\linewidth]{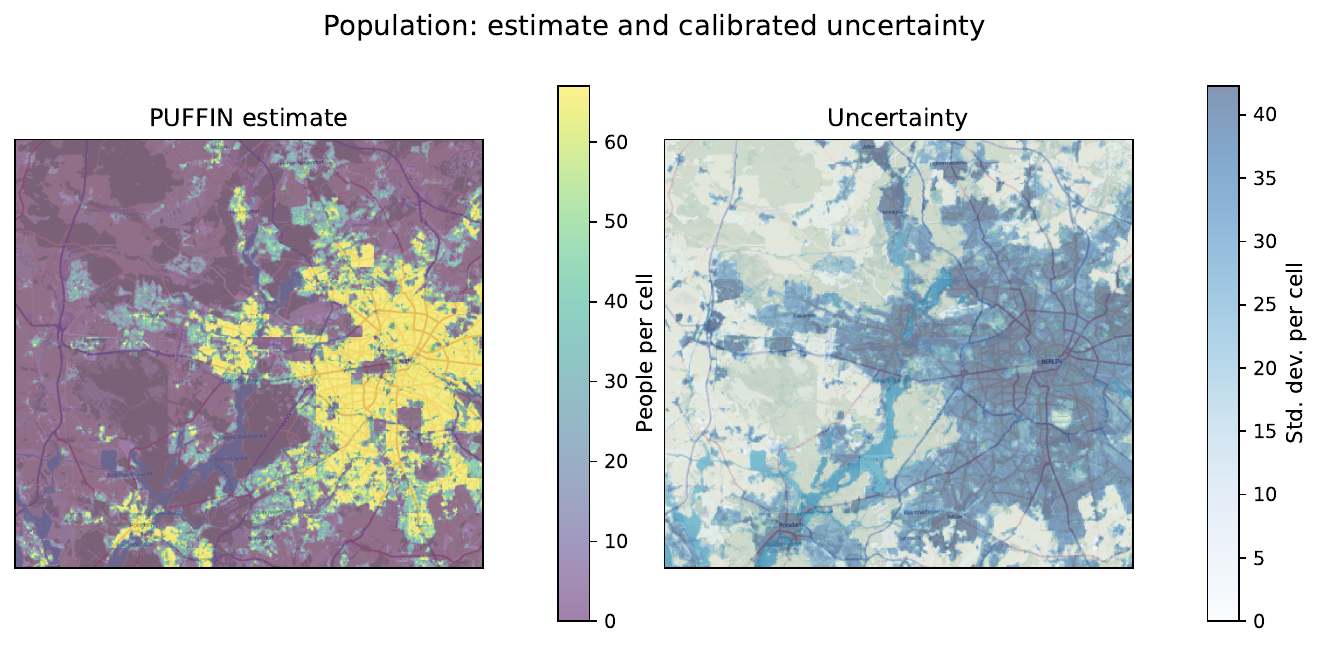}
\caption{\textbf{Calibrated per-subregion uncertainty (Berlin, $1\,$km regions split into $100\,$m subregions).} \textsc{Puffin} estimate (left) and calibrated per-subregion standard deviation (right). Uncertainty is highest across the built-up city, intermediate on water, and near zero on empty land. Gridded population products commonly provide the left map alone.}
\label{fig:uq}
\end{figure}

\paragraph{Calibration.} The per-subregion intervals reach a marginal coverage between $0.87$ and $0.93$ at the nominal $90\%$ level in every configuration, across resolutions, countries, and targets, including the negative cases below. At the finest resolutions most subregions are empty, and $\kappa$ is a quantile of validation errors, so this is a weak test, read only as evidence that the calibration is not mis-scaled. Gridded population products are usually published as one number per cell, without per-subregion intervals (Figure~\ref{fig:uq}), although hierarchical Bayesian population models \citep{leasure2020national} and recent gridded products \citep{adams2025landscan} do report them.

\begin{figure}[t]
\centering
\includegraphics[width=1.0\linewidth]{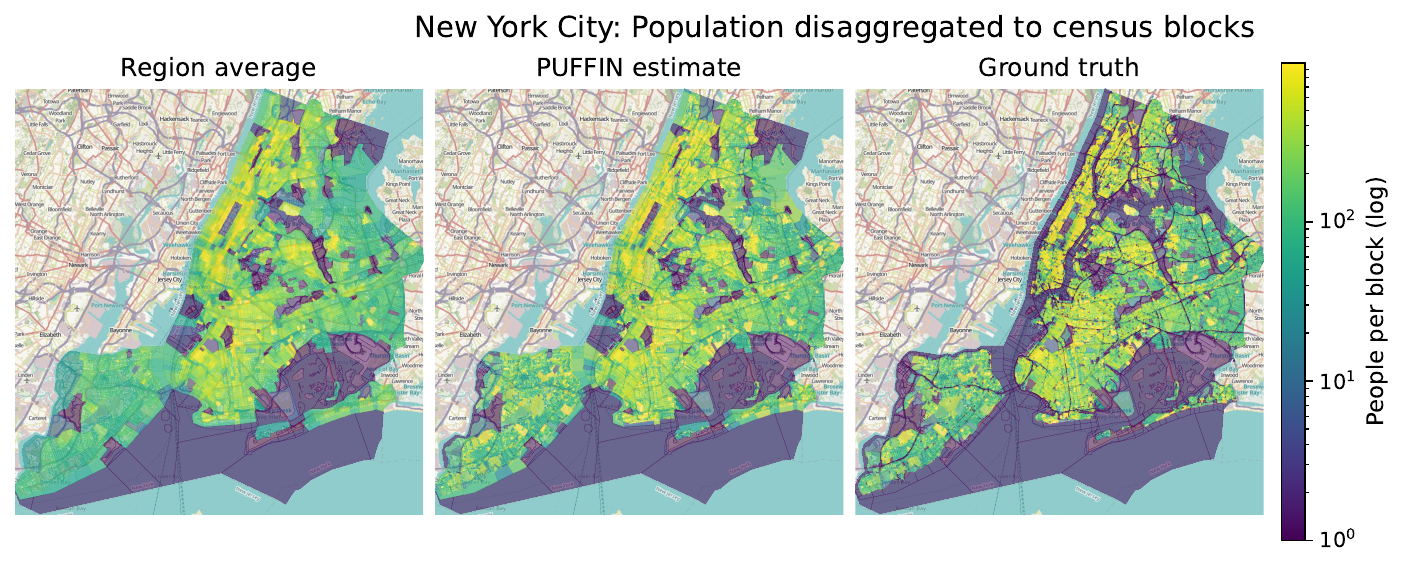}
\caption{\textbf{US population in New York City}, with census tracts as regions and census blocks as subregions. Region average (left), the \textsc{Puffin} estimate (middle), and ground truth (right). The region average gives every block of a tract the same count, the tract total divided by its number of blocks, while \textsc{Puffin} recovers the block-level structure.
}
\label{fig:usmap}
\end{figure}

\begin{table}[htbp]
\centering\small
\setlength{\tabcolsep}{4pt}
\begin{tabular}{llcccccc}
\toprule
Target & Region $\to$ subregion & Reg.-avg. & Bicubic & Goodman & \textsc{Puffin} (prior) & \textsc{Puffin} & $90\%$ cov. \\
\midrule
population & tract $\to$ block & $+0.000$ & $-0.407$ & $-1.037$ & $-0.559$ & $\mathbf{+0.377}$ & $0.90$ \\
jobs & tract $\to$ block & $+0.000$ & $+0.057$ & $+0.050$ & $-0.048$ & $\mathbf{+0.209}$ & $0.90$ \\
\midrule
turnout & county $\to$ precinct & $\mathbf{+0.000}$ & $-0.104$ & $-0.313$ & $-0.459$ & $-0.434$ & $0.90$ \\
gun violence & tract $\to$ block & $+0.000$ & $-0.079$ & $-0.075$ & $-0.089$ & $\mathbf{+0.014}$ & $0.90$ \\
\bottomrule
\end{tabular}
\caption{Within-region $R^2$ on the \textbf{US}, all subregions. The top rows are Experiment~2, with census tracts as regions and census blocks as subregions ($83{,}774$ tracts, $8.0$M blocks), where every method uses the area term. The bottom rows are Experiment~3, where precinct turnout does not follow what the embedding sees and $97\%$ of blocks record no gun-violence incident, so no method finds much to place. Best per row in bold. The baselines are deterministic, and \textsc{Puffin} is a $5$-seed mean (std $\le 0.03$) reported with its $90\%$ interval coverage. \textsc{Puffin} (prior) is the prior before conditioning; it is scored on its own split, which does not reproduce the observed total, whereas the baselines are rescaled to it.}
\label{tab:us}
\end{table}

\subsection{Experiment 3: Where \textsc{Puffin} Does Not Help}

\paragraph{Data.} For the 2024 US presidential election, we use the turnout, that is, the total number of votes cast, in each of $163{,}690$ precincts \citep{nytimes2024precinct}, which cover $3{,}096$ of the $3{,}143$ counties; Alaska and Hawaii are absent, and New England reports by township rather than by precinct. The covariates are AlphaEarth embeddings averaged over each precinct polygon. Here the regions are the $3{,}096$ counties and the subregions are their precincts. The Gun Violence Archive records geocoded incidents \citep{gunviolencearchive}. Of the $390{,}018$ geocoded incidents in the ten calendar years $2014$ to $2023$, we count the $383{,}357$ that fall inside a census block, with census tracts as regions and census blocks as subregions, as in Experiment~2.

\paragraph{Turnout.} Precinct turnout does vary within counties, where $49\%$ of its variance lies, but not in a way that follows what the embedding sees. Precincts are delineated around voters, so a dense urban precinct and a vast rural one can have similar turnout, and within a county the largest precinct is about $110\times$ the smallest by area but only about $9\times$ by turnout, taking the median over counties of the within-county maximum-to-minimum ratio. An area term therefore makes the split much worse and we use none here. Every method still adds error, and the less it moves away from the region average the better it does (Table~\ref{tab:us}), whether it redistributes by neighboring totals or by the embedding. Bicubic interpolation stays closest to the region average ($-0.10$), while Goodman-style regression ($-0.31$) and \textsc{Puffin} ($-0.43$) fall below it.

\paragraph{Gun Violence.}
Gun violence is a rare event. Only $3\%$ of census blocks record any incident, so within a tract nearly every block has the same value of zero and there is little to redistribute. \textsc{Puffin} accordingly stays close to the region average ($+0.01$), while bicubic interpolation and Goodman-style regression fall slightly below it. In both negative cases the intervals keep their nominal coverage, so the failure is in the point estimate and not hidden by a mis-scaled interval.

\section{Discussion and Limitations}

The broad lesson is that a single coarse total, read together with a satellite embedding, already pins down a large part of the fine-scale map, but only where the fine variation is real and tied to what the embedding can see.
The learned splits reach within-region $R^2$ between $0.12$ and $0.84$, above bicubic interpolation everywhere and far above it at all but the finest scale, which points to the embeddings rather than to spatial smoothness as the source of the gain. A linear density with the same embeddings keeps up on a single city and pulls ahead where training regions are few, and on irregular units both it and bicubic interpolation fall below the region average on population. Beyond accuracy, \textsc{Puffin} returns a calibrated posterior, which the baselines do not. Skill grows with the size of the subregion, which fits a hard limit at the finest scale, where one sees roofs from above but not how many people are inside.
The intervals are the one part of the method that is not label-free. The per-tile rescaling uses coarse residuals alone, but the global factor $\kappa$ is a quantile of fine-scale errors on validation subregions, which are disjoint from the subregions we score. The point estimates use no fine labels at any stage. Deploying calibrated intervals where no fine labels exist would mean transferring $\kappa$ from a dataset that has them, and we measure that it does not work (Appendix~\ref{app:details}): the per-tile stage transfers, the global factor does not, so a new resolution or unit type needs a small labeled sample of its own.

Two comparisons are missing: the same network with a squared loss and a proportional split, which would separate the probabilistic treatment from the nonlinearity, and a published gridded product. The case for the variance weighting therefore rests on the prior column and the area ablation.

The model also has structural limits. The Gaussian prior places mass on negative counts, which we remove by clipping and rescaling the mean rather than by a count likelihood; the intervals themselves are symmetric and can reach below zero, where the lower end should be read as zero. It treats subregions as independent, and it describes each subregion by a single averaged embedding, which can hide a small settlement inside a large rural unit. Its variance scales with area, which suits sparse land but understates the uncertainty of small, dense subregions, and it disaggregates one variable at a time.

\section{Related Work}

\paragraph{Population Products and Dasymetric Mapping.} WorldPop \citep{stevens2015disaggregating}, GHS-POP \citep{Schiavina_GHS-POP_R2023A_2023}, and LandScan \citep{lebakula2025landscan} split census totals using curated map layers. POMELO \citep{metzger2022fine} maps coarse counts to $100\,$m with a deep network over open geodata, and related work maps population from Sentinel imagery \citep{metzger2024high}. Tobler's mass-preserving interpolation \citep{tobler1979smooth} conserves the total but splits by a fixed rule. Deep networks that conserve mass exactly do so for climate downscaling, without a predictive distribution \citep{harder2023hard}. Regression disaggregation that reproduces the aggregate exactly goes back to \citet{chow1971best} for time series, and area-to-point kriging downscales areal data coherently and with uncertainty \citep{kyriakidis2004geostatistical}, including under non-negativity constraints \citep{yoo2006area}. Both rest on an assumed covariance structure, whereas we learn the split from covariates. Disaggregation regression fits a covariate model to aggregate counts in the same way, and is well developed for disease mapping \citep{lucas2022improving}. Closest in spirit, CLAM \citep{grossmann2026clam} likewise recovers a fine field by using the coarse total as the training signal, but targets causal local effects through a structural causal model,
whereas we disaggregate observed counts with exact conservation and calibrated uncertainty on real census and satellite data.

\paragraph{Aggregates, Embeddings, and Calibration.} Training from totals alone is an instance of learning from aggregate observations \citep{zhang2020learning,law2018variational}, specifically the count version of learning from label proportions \citep{quadrianto2008estimating}. Inferring fine-level quantities from aggregates is also the subject of ecological inference \citep{king1999solution}, whose classical regression estimator \citep{goodman1953ecological,goodman1959some} we use as a baseline. We build on AlphaEarth \citep{brown2025alphaearth}, averaging its $64$-dimensional $10\,$m embeddings over each subregion, which keeps the model small. Predicting socioeconomic outcomes from satellite imagery is well established \citep{jean2016combining,yeh2020using,rolf2021generalizable,hou2026aef}, but those methods predict a value at or within the reporting unit by supervised regression, whereas we disaggregate a coarse total below the unit under weak supervision. Our interval step is normalized split conformal prediction \citep{papadopoulos2002inductive,lei2018distribution}, rather than a quantile-regression variant \citep{romano2019conformalized}. Conformal methods usually need labels at the target resolution, whereas the step of our calibration that adapts to a new area uses coarse residuals alone, and fine labels enter only through one factor per setting, fitted on validation regions.

\section{Conclusions and Future Work}

\textsc{Puffin} casts spatial disaggregation as probabilistic constraint satisfaction. Trained only on coarse totals, it learns a prior over every subregion and conditions it on the observed totals, which gives an estimate that is non-negative, reproduces each total exactly, and carries a calibrated uncertainty. Trained from one embedding and no fine labels, it holds across resolutions, two countries, and several variables, and its failures follow a simple pattern. Where the fine-scale signal is almost absent or does not follow what the embedding sees, it falls back to or below the regional average. Needing only a published total and a globally available embedding, its point estimates apply where no fine-resolution reference grid exists, which is where disaggregation is needed most, while the intervals additionally need a small labeled sample to set their width.

\paragraph{Future Work.} The clearest next step is a head-to-head comparison against deployed gridded products \citep{stevens2015disaggregating,Schiavina_GHS-POP_R2023A_2023,lebakula2025landscan} on the same held-out regions, and an extension to more countries. The posterior also extends to a grid of cells rather than one averaged embedding per unit, to variance that grows with the predicted count rather than the area, and to several correlated variables conditioned jointly.

\subsection*{AI Use Statement}
We used generative AI tools in two ways: to aid and polish the writing, and for retrieval and discovery, that is, to cross-check our literature search for relevant work we might have overlooked. Generative AI was not used to generate the research idea or the method, and it produced no experimental result. All AI-assisted text was reviewed and edited by the authors, who take full responsibility for the content of this paper.

\subsection*{Reproducibility Statement}
Every data source is public: the Zensus 2022 grid \citep{Zensus2022}, the 2020 US census \citep{uscensus2020tiger}, LODES \citep{census_lodes}, the Gun Violence Archive \citep{gunviolencearchive}, the 2024 precinct returns \citep{nytimes2024precinct}, and the AlphaEarth embeddings \citep{brown2025alphaearth}. Section~\ref{sec:impl} states the model, the loss, the split, and the calibration, and Appendix~\ref{app:details} gives the dataset construction, the fold assignment, and every optimization setting. The code that produced every number in this paper is linked in the abstract.

\subsection*{Ethics Statement}
This work uses only publicly released data (census counts, workplace job counts, publicly reported incident records, and precinct turnout) together with public satellite embeddings. \textsc{Puffin} redistributes totals that are already published, uses no data about identified individuals, and keeps every official total unchanged. Fine-scale estimates of sensitive variables such as election turnout or violent incidents can nevertheless be mistaken for observations or misused to single out small areas, and Experiment~3 shows that such estimates can be unreliable. This is why every estimate comes with a calibrated uncertainty, and why the method is intended for planning and research at an aggregate level, not for inferring anything about individuals.

\bibliography{references}
\bibliographystyle{preprint}
\appendix

\section{Experimental Details}
\label{app:details}

\paragraph{German Grids.} The Berlin bounding box ($102\times96\,$km, $1020\times960$ cells) holds $979{,}200$ cells, of which $92{,}724$ are inhabited, and Germany holds $35.7$M cells, of which $3.1$M are inhabited. We keep every cell that has an embedding, water included, rather than masking land cover. Regions are square grids of size $R\in\{200\,\mathrm{m},1\,\mathrm{km},2\,\mathrm{km},4\,\mathrm{km},10\,\mathrm{km}\}$, each split into $2$, $5$, $10$, $20$, $50$ or $100$ subregions per side, keeping the $18$ of those whose subregions are at least $100\,$m wide. Berlin at $10\,$km is the smallest setting, with $121$ regions and therefore about $24$ test regions per fold. A subregion's outcome is the sum of the $100\,$m census cells it contains, and its covariates are the AlphaEarth embeddings \citep{brown2025alphaearth} averaged over it. We use the $2022$ AlphaEarth vintage for Germany, $2020$ for the US census data, and $2024$ for the precincts. The jobs ($2020$--$2022$) and gun-violence ($2014$--$2023$) targets have no single year, and both use the $2020$ vintage of their blocks. For the US the covariates are averaged over each block or precinct polygon instead.

\paragraph{Empty Subregions.} Nothing in the method excludes subregions that hold nobody, and the model treats them like any other subregion, so where the embedding looks like forest or water the observed totals push the predicted mean toward zero. What does not come for free is having such subregions in the data. The German Zensus grid lists only cells with at least one resident, so uninhabited cells are missing from the file rather than recorded as zero, and a model trained on the file alone could only rank cells that are already known to be inhabited. We therefore build the grid over every $100\,$m cell and label the cells the census does not list as zero. The cell-key perturbation leaves an exact zero unchanged \citep{destatis2024cellkey}, so an absent cell is a true zero rather than a rounded-down small count. Absent means zero only inside the surveyed country, so for Germany, whose bounding box reaches into neighboring countries, cells outside the national border are dropped rather than labeled zero. US census blocks need none of this, because the file already lists every block, with a count of zero where nobody lives.

\paragraph{Baselines.} Bicubic interpolation treats the regional densities $\widehat{Y}_i/\sum_j a_{i,j}$ as samples of a smooth surface and evaluates it at every subregion centroid. On regular grids we use cubic B-spline interpolation over the region centers \citep{unser1999splines}. For irregular US units, whose centers do not form a grid, we use the piecewise-cubic Clough--Tocher interpolant over the region centroids \citep{clough1965finite}. Goodman-style ecological regression is a ridge regression of the regional density on the area-weighted mean of the standardized embeddings in the region, fitted on the training regions, with the ridge strength chosen on the totals of the validation regions. In both cases the fitted density is applied to every subregion, multiplied by its area, clipped at zero, and rescaled to each regional total.

\paragraph{US Units.} The block file covers the $48$ contiguous states and the District of Columbia. Alaska and Hawaii fall outside the equal-area projection and the UTM zones in which we assemble the embeddings, so they are absent, and the $383{,}357$ gun-violence incidents we count exclude the $1{,}311$ recorded there. Within the retained states the $7{,}980{,}473$ blocks hold $325.1$M residents, about $1.3\%$ below the published total, the shortfall being blocks our embedding mosaic does not cover.

\paragraph{Calibration Details.} A tile is a $10\,$km square of the plane. At the fine resolutions it groups many regions, but once the regions themselves reach $10\,$km it holds a single one, and for counties and tracts it is the square the region's centroid falls in; where a tile holds one region, its $\gamma_t^2$ rests on a single residual and is correspondingly noisy. Stage~1 uses the totals of every region, including the regions we score. Those totals are the observed input rather than held-out labels, so this is not leakage, but the interval width is not produced out of fold. Stage~2 fits $\kappa$ on validation subregions only. The intervals are symmetric, so at low counts the lower end can fall below zero and should be read as zero.

\paragraph{Area Term Ablation.} To isolate the area factor on the prior variance we rerun the US blocks with it removed, keeping the architecture, loss, folds, epochs and seeds fixed. Within-region $R^2$ falls from $+0.365$ to $+0.260$ on population and from $+0.208$ to $+0.112$ on jobs ($3$ seeds; seed std at most $0.015$). The mean still scales with area in both arms, so the gap is attributable to the variance term alone.

\paragraph{Transferring $\kappa$.} Stage~1 needs only coarse residuals and so applies anywhere, but the global factor $\kappa$ is fitted on fine labels. We test whether it transfers by fitting $\kappa$ on one setting's validation subregions and applying it to another's test subregions, for seed $0$. The three settings differ in resolution and unit type as well as in country, so the test speaks to all three at once. Each setting calibrated on itself reaches $0.895$ to $0.900$ at the nominal $0.90$, with $\kappa=2.17$ for Germany at $10\,$km, $0.107$ for Berlin at $1\,$km and $0.677$ for the US tracts. Transferred, coverage lands between $0.32$ and $1.00$: the scale of the fine-scale residuals differs too much between a $100\,$m cell, a $1\,$km cell and a census block for one constant to serve them all. What we have measured is therefore sensitivity to the resolution and the unit, and we cannot separate out the country.

\paragraph{Optimization.} The embeddings are standardized per dimension with statistics computed on the training regions only. We train for $2{,}000$ epochs of AdamW \citep{loshchilov2019decoupled} (peak learning rate $3\cdot10^{-3}$, weight decay $10^{-4}$) with a cosine-annealed schedule, evaluating the aggregate likelihood of the validation regions each epoch and keeping the best checkpoint. The largest Germany configurations use minibatches of whole regions, where one step covers about $400{,}000$ subregions and the validation check runs every $25$ steps; every other run, including all US runs, is full-batch. Folds are fixed across seeds and methods, so \textsc{Puffin} and the baselines see exactly the same splits, and the $5$ seeds vary only the network initialization.

\section{Additional Results}
\label{app:results}

\begin{figure}[t]
\centering
\includegraphics[width=1.0\linewidth]{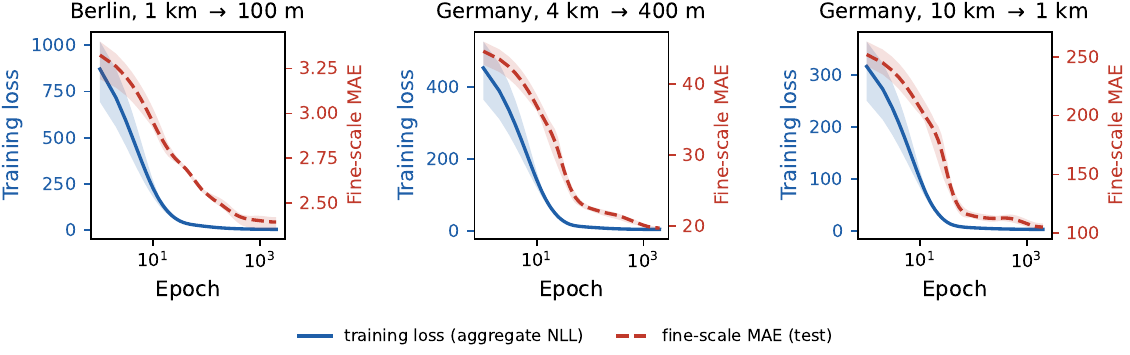}
\caption{\textbf{Training loss and fine-scale test error over training} for the remaining settings, read as Figure~\ref{fig:lc_main}: Berlin at $1\,$km, Germany at $4\,$km, and Germany at $10\,$km.}
\label{fig:lc_appendix}
\end{figure}

Interval coverage is $0.897$ to $0.902$ across the Germany configurations and $0.871$ to $0.926$ across the Berlin ones. The main paper reports within-region $R^2$ on all subregions, summarized over configurations, and Figure~\ref{fig:lc_appendix} adds the learning curves of the remaining settings. Tables~\ref{tab:berlin_within} and~\ref{tab:germany} give the per-configuration results behind Table~\ref{tab:within_main}, for every method. This appendix then collects the two other views of the same runs: the inhabited-only results and the subregion-$R^2$ results. Tables~\ref{tab:berlin_within_inhab} and~\ref{tab:germany_inhab} repeat the within-region comparison on inhabited cells only, and Table~\ref{tab:us_inhab} repeats Table~\ref{tab:us} on inhabited blocks. In those three tables the region mean of Equation~\ref{eq:within} is taken over the inhabited subregions of the region rather than over all of them, so the zero point is the inhabited-only region average. Tables~\ref{tab:berlin_cell} and~\ref{tab:us_cell} report subregion-$R^2$, which unlike within-region $R^2$ also credits the between-region differences that the observed totals give every method, so even the region average scores high. The ordering of the methods is largely unchanged, the exceptions being that bicubic interpolation and Goodman-style regression swap on US jobs, and that Goodman wins $11$ rather than $12$ Berlin configurations and $4$ rather than $2$ Germany configurations, when only inhabited subregions are scored.

\begin{figure}[t]
\centering
\includegraphics[width=1.0\linewidth]{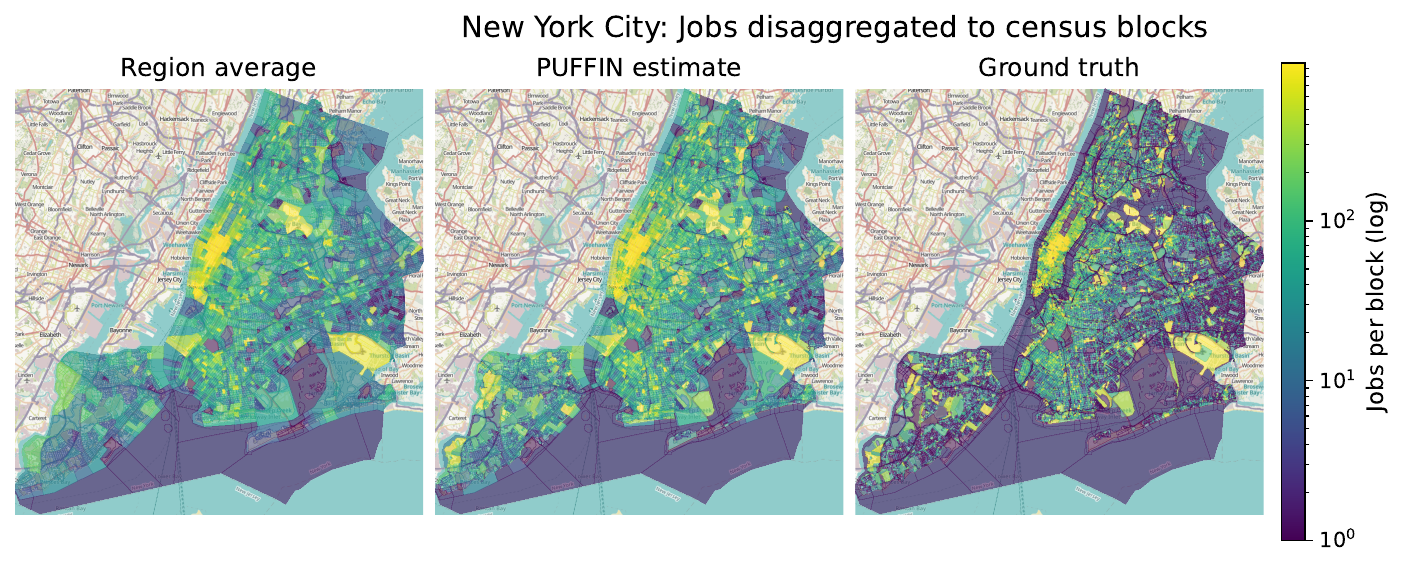}
\caption{\textbf{US jobs in New York City}, with census tracts as regions and census blocks as subregions. Region average (left), the \textsc{Puffin} estimate (middle), and ground truth (right). Jobs concentrate in a few small commercial blocks, which the region average cannot express and \textsc{Puffin} recovers from the embeddings and the tract totals alone.}
\label{fig:usjobs}
\end{figure}

\begin{table}[H]
\centering\small
\setlength{\tabcolsep}{4pt}
\begin{tabular}{llcccccc}
\toprule
& & \multicolumn{6}{c}{Subregions per side ($R/s$)} \\
\cmidrule(lr){3-8}
Method & Region size $R$ & $2$ & $5$ & $10$ & $20$ & $50$ & $100$ \\
\midrule
\multirow{5}{*}{Bicubic} & $200\,$m & .11 & -- & -- & -- & -- & -- \\
 & $1\,$km & .21 & .11 & .06 & -- & -- & -- \\
 & $2\,$km & .17 & .11 & .07 & .04 & -- & -- \\
 & $4\,$km & .20 & .11 & .07 & .05 & -- & -- \\
 & $10\,$km & .54 & .27 & .18 & .12 & .08 & .05 \\
\midrule
\multirow{5}{*}{Goodman} & $200\,$m & \textbf{.17} & -- & -- & -- & -- & -- \\
 & $1\,$km & \textbf{.49} & \textbf{.40} & \textbf{.28} & -- & -- & -- \\
 & $2\,$km & \textbf{.53} & .46 & .38 & .28 & -- & -- \\
 & $4\,$km & \textbf{.57} & .50 & .44 & .36 & -- & -- \\
 & $10\,$km & \textbf{.71} & \textbf{.70} & \textbf{.59} & \textbf{.50} & \textbf{.39} & \textbf{.29} \\
\midrule
\multirow{5}{*}{\textsc{Puffin} (prior)} & $200\,$m & .05 & -- & -- & -- & -- & -- \\
 & $1\,$km & .25 & .30 & .21 & -- & -- & -- \\
 & $2\,$km & .37 & .41 & .34 & .24 & -- & -- \\
 & $4\,$km & .44 & .44 & .40 & .32 & -- & -- \\
 & $10\,$km & .56 & .50 & .43 & .37 & .30 & .21 \\
\midrule
\multirow{5}{*}{\textsc{Puffin}} & $200\,$m & .12 & -- & -- & -- & -- & -- \\
 & $1\,$km & .41 & .38 & .28 & -- & -- & -- \\
 & $2\,$km & .49 & \textbf{.48} & \textbf{.42} & \textbf{.30} & -- & -- \\
 & $4\,$km & .54 & \textbf{.51} & \textbf{.47} & \textbf{.39} & -- & -- \\
 & $10\,$km & .68 & .58 & .52 & .45 & .38 & .28 \\
\bottomrule
\end{tabular}
\caption{Within-region $R^2$ on \textbf{Berlin}, all $100\,$m cells, per configuration and method. The \textsc{Puffin} rows are those of Table~\ref{tab:within_main}. Each row is a coarse region size $R$ and each column the number of subregions per side, so a subregion is $R$ divided by that number ($100$ means $100{\times}100$ subregions), and blanks are configurations with sub-$100\,$m subregions. The region-average baseline scores $0$ by construction. Best method per configuration in bold, compared on the unrounded values. \textsc{Puffin}: $5$-seed mean (std $\le 0.024$), and the baselines are deterministic. \textsc{Puffin} (prior) is the prior alone, before conditioning on the totals.}
\label{tab:berlin_within}
\end{table}

\begin{table}[H]
\centering\small
\setlength{\tabcolsep}{4pt}
\begin{tabular}{llcccccc}
\toprule
& & \multicolumn{6}{c}{Subregions per side ($R/s$)} \\
\cmidrule(lr){3-8}
Method & Region size $R$ & $2$ & $5$ & $10$ & $20$ & $50$ & $100$ \\
\midrule
\multirow{5}{*}{Bicubic} & $200\,$m & .14 & -- & -- & -- & -- & -- \\
 & $1\,$km & .23 & .13 & .08 & -- & -- & -- \\
 & $2\,$km & .16 & .10 & .07 & .05 & -- & -- \\
 & $4\,$km & .16 & .09 & .06 & .04 & -- & -- \\
 & $10\,$km & .22 & .12 & .07 & .05 & .03 & .02 \\
\midrule
\multirow{5}{*}{Goodman} & $200\,$m & \textbf{.19} & -- & -- & -- & -- & -- \\
 & $1\,$km & \textbf{.62} & .48 & .34 & -- & -- & -- \\
 & $2\,$km & .69 & .59 & .49 & .36 & -- & -- \\
 & $4\,$km & .73 & .65 & .58 & .48 & -- & -- \\
 & $10\,$km & .80 & .74 & .66 & .59 & .47 & .35 \\
\midrule
\multirow{5}{*}{\textsc{Puffin} (prior)} & $200\,$m & .09 & -- & -- & -- & -- & -- \\
 & $1\,$km & .43 & .39 & .28 & -- & -- & -- \\
 & $2\,$km & .59 & .53 & .44 & .32 & -- & -- \\
 & $4\,$km & .67 & .63 & .55 & .44 & -- & -- \\
 & $10\,$km & .71 & .68 & .65 & .56 & .43 & .32 \\
\midrule
\multirow{5}{*}{\textsc{Puffin}} & $200\,$m & .18 & -- & -- & -- & -- & -- \\
 & $1\,$km & .58 & \textbf{.48} & \textbf{.35} & -- & -- & -- \\
 & $2\,$km & \textbf{.70} & \textbf{.61} & \textbf{.51} & \textbf{.37} & -- & -- \\
 & $4\,$km & \textbf{.77} & \textbf{.69} & \textbf{.61} & \textbf{.50} & -- & -- \\
 & $10\,$km & \textbf{.84} & \textbf{.75} & \textbf{.70} & \textbf{.62} & \textbf{.48} & \textbf{.36} \\
\bottomrule
\end{tabular}
\caption{Within-region $R^2$ on \textbf{Germany}, all $35.7$M cells, per configuration and method, read as Table~\ref{tab:berlin_within}. \textsc{Puffin}: $5$-seed mean (std $\le 0.016$), and the baselines are deterministic. \textsc{Puffin} (prior) is the prior alone, before conditioning on the totals.}
\label{tab:germany}
\end{table}

\begin{table}[H]
\centering\small
\setlength{\tabcolsep}{3pt}
\begin{tabular}{lcccccccc}
\toprule
& \multicolumn{4}{c}{inhabited} & \multicolumn{4}{c}{all $100\,$m cells} \\
\cmidrule(lr){2-5}\cmidrule(lr){6-9}
Config ($R$ / $s$) & Reg.-avg. & Bicubic & Goodman & \textsc{Puffin} & Reg.-avg. & Bicubic & Goodman & \textsc{Puffin} \\
\midrule
$10\,$km / $100\,$m & 0.103 & 0.168 & \textbf{0.409} & 0.401 & 0.247 & 0.287 & \textbf{0.466} & 0.460 \\
$2\,$km / $100\,$m & 0.310 & 0.339 & 0.499 & \textbf{0.511} & 0.385 & 0.411 & 0.557 & \textbf{0.569} \\
$1\,$km / $100\,$m & 0.410 & 0.444 & \textbf{0.557} & \textbf{0.557} & 0.468 & 0.501 & \textbf{0.619} & 0.617 \\
$1\,$km / $200\,$m & 0.581 & 0.621 & \textbf{0.734} & 0.727 & 0.622 & 0.663 & \textbf{0.775} & 0.767 \\
$10\,$km / $1\,$km & 0.506 & 0.594 & \textbf{0.791} & 0.752 & 0.527 & 0.611 & \textbf{0.805} & 0.771 \\
$10\,$km / $5\,$km & 0.828 & 0.920 & \textbf{0.950} & 0.945 & 0.829 & 0.920 & \textbf{0.950} & 0.946 \\
\bottomrule
\end{tabular}
\caption{Subregion-$R^2$ on \textbf{Berlin} for representative configurations, inhabited cells and all $100\,$m cells. Unlike within-region $R^2$, this metric also credits the between-region differences that the observed totals give every method, so even the region average scores high. Best per group in bold, and \textsc{Puffin} std $\le 0.01$.}
\label{tab:berlin_cell}
\end{table}

\begin{table}[H]
\centering\small
\setlength{\tabcolsep}{3pt}
\begin{tabular}{lcccc}
\toprule
Target & Reg.-avg. & Bicubic & Goodman & \textsc{Puffin} \\
\midrule
population & $0.222$ & $-0.095$ & $-0.584$ & $\mathbf{0.515}$ \\
jobs & $0.087$ & $0.139$ & $0.133$ & $\mathbf{0.277}$ \\
\bottomrule
\end{tabular}
\caption{Subregion-$R^2$ on the \textbf{US}, census tracts as regions and census blocks as subregions, all blocks, for the runs of Table~\ref{tab:us}. Best per row in bold, and \textsc{Puffin} std $\le 0.02$.}
\label{tab:us_cell}
\end{table}

\begin{table}[H]
\centering\small
\setlength{\tabcolsep}{4pt}
\begin{tabular}{llcccccc}
\toprule
& & \multicolumn{6}{c}{Subregions per side ($R/s$)} \\
\cmidrule(lr){3-8}
Method & Region size $R$ & $2$ & $5$ & $10$ & $20$ & $50$ & $100$ \\
\midrule
\multirow{5}{*}{Bicubic} & $200\,$m & .06 & -- & -- & -- & -- & -- \\
 & $1\,$km & .19 & .07 & .03 & -- & -- & -- \\
 & $2\,$km & .16 & .09 & .05 & .03 & -- & -- \\
 & $4\,$km & .19 & .11 & .07 & .04 & -- & -- \\
 & $10\,$km & .54 & .27 & .18 & .13 & .09 & .07 \\
\midrule
\multirow{5}{*}{Goodman} & $200\,$m & \textbf{.10} & -- & -- & -- & -- & -- \\
 & $1\,$km & \textbf{.47} & \textbf{.33} & .16 & -- & -- & -- \\
 & $2\,$km & \textbf{.52} & .43 & .32 & .18 & -- & -- \\
 & $4\,$km & \textbf{.57} & .48 & .41 & .31 & -- & -- \\
 & $10\,$km & \textbf{.71} & \textbf{.69} & \textbf{.57} & \textbf{.47} & \textbf{.37} & \textbf{.26} \\
\midrule
\multirow{5}{*}{\textsc{Puffin} (prior)} & $200\,$m & .05 & -- & -- & -- & -- & -- \\
 & $1\,$km & .26 & .27 & .14 & -- & -- & -- \\
 & $2\,$km & .37 & .40 & .31 & .17 & -- & -- \\
 & $4\,$km & .44 & .44 & .40 & .31 & -- & -- \\
 & $10\,$km & .56 & .50 & .43 & .36 & .32 & .24 \\
\midrule
\multirow{5}{*}{\textsc{Puffin}} & $200\,$m & .09 & -- & -- & -- & -- & -- \\
 & $1\,$km & .39 & .31 & \textbf{.16} & -- & -- & -- \\
 & $2\,$km & .48 & \textbf{.44} & \textbf{.34} & \textbf{.19} & -- & -- \\
 & $4\,$km & .53 & \textbf{.49} & \textbf{.43} & \textbf{.33} & -- & -- \\
 & $10\,$km & .68 & .58 & .49 & .42 & .35 & .25 \\
\bottomrule
\end{tabular}
\caption{Within-region $R^2$ on \textbf{Berlin}, inhabited cells only, read as Table~\ref{tab:berlin_within}. \textsc{Puffin}: $5$-seed mean (std $\le 0.024$), and the baselines are deterministic. \textsc{Puffin} (prior) is the prior alone, before conditioning on the totals.}
\label{tab:berlin_within_inhab}
\end{table}

\begin{table}[H]
\centering\small
\setlength{\tabcolsep}{4pt}
\begin{tabular}{llcccccc}
\toprule
& & \multicolumn{6}{c}{Subregions per side ($R/s$)} \\
\cmidrule(lr){3-8}
Method & Region size $R$ & $2$ & $5$ & $10$ & $20$ & $50$ & $100$ \\
\midrule
\multirow{5}{*}{Bicubic} & $200\,$m & .09 & -- & -- & -- & -- & -- \\
 & $1\,$km & .20 & .08 & .03 & -- & -- & -- \\
 & $2\,$km & .16 & .08 & .05 & .03 & -- & -- \\
 & $4\,$km & .16 & .08 & .05 & .03 & -- & -- \\
 & $10\,$km & .22 & .12 & .07 & .05 & .03 & .03 \\
\midrule
\multirow{5}{*}{Goodman} & $200\,$m & \textbf{.11} & -- & -- & -- & -- & -- \\
 & $1\,$km & \textbf{.59} & \textbf{.37} & \textbf{.16} & -- & -- & -- \\
 & $2\,$km & .67 & .53 & .37 & .18 & -- & -- \\
 & $4\,$km & .73 & .63 & .52 & .37 & -- & -- \\
 & $10\,$km & .80 & .73 & .65 & .54 & .36 & .20 \\
\midrule
\multirow{5}{*}{\textsc{Puffin} (prior)} & $200\,$m & .08 & -- & -- & -- & -- & -- \\
 & $1\,$km & .42 & .31 & .14 & -- & -- & -- \\
 & $2\,$km & .59 & .48 & .35 & .16 & -- & -- \\
 & $4\,$km & .68 & .61 & .51 & .36 & -- & -- \\
 & $10\,$km & .72 & .68 & .65 & .54 & .37 & .20 \\
\midrule
\multirow{5}{*}{\textsc{Puffin}} & $200\,$m & .11 & -- & -- & -- & -- & -- \\
 & $1\,$km & .54 & .37 & .16 & -- & -- & -- \\
 & $2\,$km & \textbf{.69} & \textbf{.54} & \textbf{.39} & \textbf{.18} & -- & -- \\
 & $4\,$km & \textbf{.77} & \textbf{.66} & \textbf{.55} & \textbf{.39} & -- & -- \\
 & $10\,$km & \textbf{.84} & \textbf{.74} & \textbf{.69} & \textbf{.58} & \textbf{.40} & \textbf{.22} \\
\bottomrule
\end{tabular}
\caption{Within-region $R^2$ on \textbf{Germany}, inhabited cells only, read as Table~\ref{tab:berlin_within}. \textsc{Puffin}: $5$-seed mean (std $\le 0.017$), and the baselines are deterministic. \textsc{Puffin} (prior) is the prior alone, before conditioning on the totals.}
\label{tab:germany_inhab}
\end{table}

\begin{table}[H]
\centering\small
\setlength{\tabcolsep}{3pt}
\begin{tabular}{lcccccccc}
\toprule
& \multicolumn{4}{c}{within-region $R^2$} & \multicolumn{4}{c}{subregion-$R^2$} \\
\cmidrule(lr){2-5}\cmidrule(lr){6-9}
Target & Reg.-avg. & Bicubic & Goodman & \textsc{Puffin} & Reg.-avg. & Bicubic & Goodman & \textsc{Puffin} \\
\midrule
population & $+0.000$ & $-0.297$ & $-0.901$ & $\mathbf{+0.429}$ & $0.231$ & $0.035$ & $-0.406$ & $\mathbf{0.568}$ \\
jobs & $+0.000$ & $+0.068$ & $+0.073$ & $\mathbf{+0.183}$ & $0.094$ & $0.174$ & $0.179$ & $\mathbf{0.278}$ \\
\bottomrule
\end{tabular}
\caption{\textbf{US}, census tracts as regions and census blocks as subregions, inhabited blocks only, read as Table~\ref{tab:us}. \textsc{Puffin} std $\le 0.02$.}
\label{tab:us_inhab}
\end{table}

\end{document}